\documentclass[11pt]{article}

\usepackage[final]{acl}

\usepackage{times}
\usepackage{latexsym}

\usepackage[T1]{fontenc}
\usepackage{afterpage}
\usepackage{tabularx}
\usepackage{enumerate}
\usepackage[utf8]{inputenc}

\usepackage{microtype}

\usepackage{inconsolata}

\usepackage{graphicx}

\title{Answering Questions in Stages: Prompt Chaining for Contract QA}

\author{Adam Roegiest \\
  Zuva Inc.\\ Toronto, Ontario, Canada\\
  \texttt{adam@zuva.ai } \\\And
  Radha Chitta \\
  Zuva Inc.\\ Toronto, Ontario, Canada\\
  \texttt{radha.chitta@zuva.ai} \\}

\begin{document}
\newcommand{\red}[1]{\textcolor{red}{#1}}
\maketitle
\begin{abstract}
Finding answers to legal questions about clauses in contracts is an important form of analysis in many legal workflows (e.g., understanding market trends, due diligence, risk mitigation) but more important is being able to do this at scale. Prior work showed that it is possible to use large language models with simple zero-shot prompts to generate structured answers to questions, which can later be incorporated into legal workflows. Such prompts, while effective on simple and straightforward clauses, fail to perform when the clauses are long and contain information not relevant to the question. In this paper, we propose two-stage prompt chaining to produce structured answers to multiple-choice and multiple-select questions and show that they are more effective than simple prompts on more nuanced legal text. We analyze situations where this technique works well and areas where further refinement is needed, especially when the underlying linguistic variations are more than can be captured by simply specifying possible answers. 
Finally, we discuss future research that seeks to refine this work by improving stage one results by making them more question-specific.
\end{abstract}

\section{Introduction}
As generative AI has slowly found its way into many different products and research endeavors, there are continual attempts to determine the best ways to leverage this technology and where the limits lie during adoption. Recent developments in natural language generation and understanding have led to several conversational generative systems that allow users to ``chat'' with their contracts and find answers to questions (e.g., ChatPDF\footnote{\url{https://www.chatpdf.com/}}, docGPT\footnote{\url{https://github.com/cesarhuret/docGPT}}, CaseText\footnote{\url{https://casetext.com/}}). Such systems seek to empower users to more easily understand the contents of their documents and improve decision making. This is particularly important in the case of legal document workflows (e.g., contract drafting, negotiation, mergers and acquisitions due diligence) where mistakes or a lack of understanding of one or more legal documents can have an out-sized effect (e.g., over-valuing a company~\cite{autonomy}). However, the conversational style of these systems can make it difficult to operationalize outputs when scaling to large document collections. While convenient on a per-document basis, conversational aspects can result in less actionable outcomes at scale when trying to perform more complex forms of analysis (e.g., measuring different risks across contracts). Generating structured answers (such as multiple choice and multiple select) can provide a more effective and scalable mechanism to allow legal professionals to utilize these advancements in generative AI to more efficiently triage and review documents as well as build automated workflows and data pipelines.

In past work~\cite{roegiest2023questions,roegiest2023}, we demonstrated that it is possible to build prompts to generate structured answers to questions based on clauses extracted from contracts. We also built reusable prompt templates to minimize the effort involved in tailoring prompts to the clause, question, and answer combinations. The users could instead focus on refining the answer options to yield accurate answers.
This reflects a growing trend to use prompt templates to consistently solve similar classes of problems~\cite{schmidt2023towards,ye2024prompt}. While these prompt templates performed well on simple clauses ( e.g., environmental indemnity, non-solicitation), further experimentation revealed issues in their ability to handle long and multi-faceted clauses containing information superfluous to the question but responsive to the underlying legal concept (e.g., change of control, types of insurance).

Using extremely specific answer phrasing and additional bespoke information in prompts tends to help address long and complex clauses, but the added benefit was never particularly large and required a great deal of trial and effort to get ``just the right'' combination, which was effort that we sought to avoid by using prompt templates. We conceived of a two-stage chain of prompts strategy to tackle this problem. In the first stage, the generative model would be asked to provide a summary of the relevant legal text in the context of the question. The second stage would then use this summary in place of the legal text in our original~\cite{roegiest2023questions,roegiest2023} prompt template. The underlying intuition is that, in the summary generation stage, the model could ``distill'' the legal text to only the concepts relevant to the question and would facilitate better mapping to relevant answer options.
Our experiments show that this technique is more effective than our prior attempts in most cases. With minor modifications that allowed more structured, concept-specific restrictions as well as including the answer options in the summary generation stage, we are able to achieve improved question-answering abilities. We also investigated cases where this technique is not as effective as expected and attempt to identify the scenarios where this technique is most effective and those where it falls short.
 
\section{Background}
\subsection{Large Language Models}
Large language models are self-supervised generative models that are trained to predict the next token in a sequence of tokens. While the early simple language models such as BERT~\cite{devlin2018bert} and T5~\cite{raffel2020exploring} had comparatively few parameters, recent large language models such as the GPT-n series\footnote{https://platform.openai.com/docs/models}, the Llama-n series~\cite{touvron2023llama,llama2}, Gemma~\cite{gemmateam2024gemma}, Mistral~\cite{jiang2023mistral} etc. contain billions of trainable parameters. They can emulate complex reasoning capabilities~\cite{huang-chang-2023-towards} and have performed well in many natural language tasks including question-answering, summarization, and translation~\cite{sanh2021multitask, wang2022benchmarking, minaee2024large}. 
\subsection{Prompt Engineering}
Prompt engineering is the range of techniques used to structure the instructions that can be interpreted and understood by the large language model. Building the right prompt is a challenging task as minor variations in a prompt can lead to substantial differences in the model's responses and can lead to hallucinations or false/irrelevant information~\cite{ji2023survey,magesh2024hallucinationfree}. There has been extensive research in this area over the past couple of years, with the main aim of making it easier for LLM users to obtain relevant and accurate results~\cite{sahoo2024systematic, chen2023unleashing}. Several strategies including automatic prompt optimization~\cite{cao2023beautifulprompt, hao2024optimizing,ling2023towards}, meta-prompting~\cite{suzgun2024metaprompting, ye2024prompt}, interactive prompt refinement~\cite{feng2024promptmagician}, and prompt templating~\cite{schmidt2023towards,roegiest2023,roegiest2023questions} have been developed to reduce hallucinations and make prompt tuning efficient. 
\newline
\newline
\textbf{Chain-Of-Thought (``CoT'') Prompting}~\cite{wei2023chainofthought} involves guiding a model step-by-step toward the answer using two or more prompts. Typical zero-shot CoT prompts include instructions that ask the large language model to ``generate reasoning'' or ``describe its thought process'' as it produces the final answer. This leads the LLM to not only generate an end result, but also detail the series of intermediate steps that potentially led to that result. By generating these intermediate steps, LLMs can produce better answers than they would directly~\cite{prystawski2023reasoning}.
\newline
\newline
\textbf{Prompt Chaining}~\cite{promptchaining}, inspired by chain-of-thought prompting, involves splitting the task into multiple sub-tasks and using the response of the LLM from one task as input to the next. Prompt chaining is useful to accomplish complex tasks that an LLM might struggle to address when faced with a very complex prompt. For instance, prompt chaining was found to produce more accurate and refined text summaries than single stage prompts~\cite{sun2024promptchainingstepwiseprompt}.

\subsection{Legal and LLMs}
In the legal domain, LLMs have been primarily used for chat-style interactions~\cite{kuppa2023chain}, case retrieval~\cite{li2023sailer}, case summarization~\cite{nay2023large}, and simple yes/no questions for predicting legal judgments~\cite{trautmann2022legal, yu2022legal}. The effects of hallucination and mis-attribution have been explored in the context of production legal systems~\cite{magesh2024hallucinationfree}. In our previous work~\cite{roegiest2023,roegiest2023questions}, we explored the use of LLMs for generating structured outputs from contracts, and the creation of simple prompt templates to generate such structured outputs. To the best of our knowledge, this article is one of the first to investigate the application of prompt chaining to the generation of structured outputs from contracts.    

\section{Methodology}
\begin{figure}
\centering
\fontsize{5}{6}\selectfont
\begin{tabularx}{\columnwidth}{p{1.5cm} | X}
Question & Options \\\hline
Q1: What are the obligations and requirements for a Change of Control? & 
\begin{enumerate}[a)]
\vspace{-0.4cm}\item One or more parties must give notice of a change of control, either prior to or after the change of control occurs.
\vspace{-0.3cm}\item One or more parties requires consent, approval or mutual agreement in order to undertake a change of control.
\vspace{-0.3cm}\item One or more parties has obligation(s) prior to or after a change of control occurs other than notice, consent or approval. These could include certain conditions in order to undertake a change of control (including restrictions on who can take control, such as competitors or affiliates, rights of first refusal or options to purchase), executing documents or providing certain documentation or payment of a transfer fee, and similar requirements/conditions. 
\vspace{-0.3cm}\item One or more parties is not permitted to undertake a change of control.
\vspace{-0.3cm}\item The agreement may be terminated if a change of control occurs, either at a party’s option or automatically. Payment of a termination fee may be required.
\vspace{-0.3cm}\item The agreement remains in effect regardless of a change of control.
\vspace{-0.3cm}\item The provided text lacks sufficient information to determine any ramifications of a change of control.
\vspace{-0.2cm}
\end{enumerate} \\
\hline
Q2: What are the obligations of the parties for assigning the agreement? & 
\begin{enumerate}[a)]
\vspace{-0.4cm}\item A party may assign the agreement freely.
\vspace{-0.3cm}\item A party may assign the agreement if it provides notice.
\vspace{-0.3cm}\item A party may not assign the agreement unless consent or approval is obtained. 
\vspace{-0.3cm}\item  If a party assigns the agreement, the agreement can be terminated.
\vspace{-0.3cm}\item A party may assign the agreement but must pay a termination fee.
\vspace{-0.3cm}\item A party may not assign the agreement.
\vspace{-0.2cm}
\end{enumerate} \\
\hline
Q3:  What type(s) of insurance is a party required to have? & 
\begin{enumerate}[a)]
\vspace{-0.4cm}\item Errors and Omissions/Professional Liability Insurance.
\vspace{-0.3cm}\item Umbrella/Excess Liability Insurance.
\vspace{-0.3cm}\item Comprehensive Automobile/Motor Vehicle Liability Insurance. 
\vspace{-0.3cm}\item Workers'/Workmans' Compensation Insurance.
\vspace{-0.3cm}\item Employer's Liability Insurance.
\vspace{-0.3cm}\item General Liability/Comprehensive General/Public Liability Insurance or similar terms.
\vspace{-0.3cm}\item Product Liability Insurance.
\vspace{-0.3cm}\item "All Risk"/Property/Physical Damage Insurance.
\vspace{-0.3cm}\item Other type(s) of insurance that are not listed in the above options.
\vspace{-0.2cm}
\end{enumerate} \\
\hline
Q4: What are the force majeure trigger events? & 
\begin{enumerate}[a)]
\vspace{-0.4cm}\item Acts of God or natural disasters. 
\vspace{-0.3cm}\item Wars or government actions.
\vspace{-0.3cm}\item Public health events. 
\vspace{-0.3cm}\item Utility or infrastructure failures.
\vspace{-0.3cm}\item Labour disruptions or third-party actions.
\vspace{-0.3cm}\item There is catch-all language in the clause.
\vspace{-0.2cm}
\end{enumerate} \\
\hline
\end{tabularx}
\caption{Questions used for testing the LLM models}
\label{fig:questions}
\vspace{-3em}
\end{figure}

In exploring how best to develop prompting strategies to effectively and efficiently utilize generative AI (i.e., being cost-efficient), we focus on this work on four questions that we have found to be most difficult to elicit consistent and effective performance. The questions are interesting as their associated legal text (i.e., one or more clauses) tend to be long and contain information beyond what is required to answer the question (e.g., conditions of a change of control, value of insurance). In our previous work~\cite{roegiest2023}, we explored over 700 prompt and clause combinations following several prompt styles and built prompt templates which worked well on a number of questions based on legal clauses. These templates, however, performed poorly for these questions despite many attempts at improving answer formulation or adding additional customized information to the prompt and would require many highly tailored modifications of the existing templates to achieve anything approaching usability. We use the following questions, with answer options provided in Figure~\ref{fig:questions}:
\begin{itemize}
 \setlength\itemsep{-0.3em}
    \item What are the parties' obligations when there is a change of control? 
    \item What are the obligations of the parties for assigning the agreement? 
    \item What types of insurance must the parties possess? 
    \item What events would trigger the force majeure clause? 
\end{itemize}

The clauses corresponding to these questions: change of control\footnote{Change of Control clauses outline the treatment of a party's rights (such as consent, payment or termination) in connection with a change in ownership or management of the other party to the agreement}, assignment\footnote{Assignment clauses outline a party's rights to engage in a transfer of ownership or assign their contractual obligations and rights to a different contracting party.}, insurance\footnote{An insurance clause establishes what insurance one or more parties must procure in connection with an agreement.}, and force majeure\footnote{A force majeure clause establishes what happens if a breach of contract occurs due to an unforeseeable event (a force majeure event).} are commonly found in contracts between companies. We used a collection of previously annotated legal documents, annotated similarly to those described in
\cite{roegiest2018}, collected from EDGAR\footnote{\url{https://www.sec.gov/edgar}} and SEDAR\footnote{\url{https://www.sedarplus.ca/}} document repositories. We note that the documents were initially annotated to train machine learning systems to identify the relevant concepts and that the annotations may not always align exactly with the purposes of the questions (e.g., they may include information superfluous to the question but relevant to the underlying concept). Our in-house lawyers reviewed about 200 annotated clauses and provided the correct answer(s) for each question\footnote{This dataset will be released later for public use.}.

Our experiments are broken down into two types of experiments: single stage prompts where a generative model matches answers to clauses directly; and, two stage prompt chains where a model first produces a question-specific summary of the legal text and a second model maps that summary to answer options. In all of our two-stage experiments, the same large language model is used for both stages but not as part of a ``conversation'' (i.e., two distinct interactions are made with the model). We do this to isolate any effects that could occur by utilizing a conversational context as well as reduce context length to improve the ability to use smaller models in the future as well as reduce costs. We used OpenAI's 
GPT-4o and GPT-4-Turbo models for all experiments. All model hyperparameters were set to the default values and the \emph{temperature} was set to zero for all experiments to minimize any inconsistency during generation.

Model responses are scored with respect to per-answer precision and recall, and ``exact match'' accuracy (i.e., how frequently the model selects only and exactly the correct answers). We find these metrics allow us to understand how well the model selects answers both individually and overall without relying on complex scoring regimes.

\section{Results}
\subsection{Single Stage Prompts}
\begin{table}[p]
\scriptsize
  \centering
\resizebox{\columnwidth}{!}{
  \begin{tabular}{|c|c||c|c|c|c||c|c|c|c|}
  \hline
  \textbf{Prompt}  & \textbf{Metric} & \multicolumn{4}{c||}{\textbf{GPT-4o}} & \multicolumn{4}{c|}{\textbf{GPT-4 Turbo}} \\
    \hline   
     & & \textbf{Q1} & \textbf{Q2} & \textbf{Q3}& \textbf{Q4} &\textbf{Q1} & \textbf{Q2} & \textbf{Q3}& \textbf{Q4} \\
    \hline   
    
   \textbf{P1}  & Exact Match & 0.47 &0.84& 0.51&0.68 &0.55 &0.81 &0.51&0.73\\
    \cline{2-10}
    & Precision& 0.65 & 0.84& 0.86&0.95& 0.72& 0.79&0.85& 0.96\\
    \cline{2-10}
     & Recall& 0.89& 0.67& 0.88&0.91&0.80 & 0.86& 0.87&0.94\\
     \cline{2-10}
    \hline
    \hline
\textbf{P2}  & Exact Match&0.49  &0.83 &0.48&0.63&0.41 & 0.79 & 0.49 &0.73\\
    \cline{2-10}
    & Precision& 0.62& 0.83&0.86& 0.99&0.67 &0.78 &0.84 &0.96\\
    \cline{2-10}
     & Recall& 0.82& 0.72 &0.83& 0.86&0.77 & 0.73 &0.84& 0.92\\
     \cline{2-10}
    \hline
    \hline
\textbf{P3}  & Exact Match & 0.51 &0.73&  0.51&0.54&0.50 &0.84 &0.44&0.57\\
    \cline{2-10}
    & Precision&0.69 & 0.72& 0.88&0.96& 0.71&0.83 & 0.87&0.96\\
    \cline{2-10}
     & Recall&0.87 & 0.83& 0.84&0.89& 0.87&0.79 &0.82 &0.89\\
     \cline{2-10}
    \hline
    \hline
\textbf{P4}  & Exact Match & 0.57 &0.83& 0.58&0.68&0.66 &0.80 & 0.54&0.73\\
    \cline{2-10}
    & Precision&0.69 & 0.77& 0.88&0.95& 0.78& 0.72& 0.87&0.97\\
    \cline{2-10}
     & Recall&0.86 & 0.82&  0.89&0.93& 0.84&0.91 &0.87&0.92\\
     \cline{2-10}
    \hline
  \end{tabular}}
  \caption{Exact match accuracy, average Precision, and average Recall for GPT-4o and GPT-4 Turbo on four legal questions using different prompts for structured answer generation with a fixed set of options.}
  \label{tbl:results}
  \vspace{-2em}
\end{table}
Having previously sought to find a unifying set of template prompts~\cite{roegiest2023questions}, we turned to a single stage prompt that directly examines legal text and produces one or more relevant answers (Figure~\ref{fig:single-prompts}). 
This style of prompt has worked reasonably well on many legal clause-based questions and not required exhaustive re-framing of answer options. However, for the questions of interest in this work, the prompt performs quite poorly as seen in Table~\ref{tbl:results}. Except for question \textbf{Q2}, this prompt's exact match accuracy is less than 80\% and is close to 50\% for questions \textbf{Q1} and \textbf{Q3}. Given that each of the questions has 6 or more answer options, an exact match accuracy close to 50\%  (i.e., the prompt selects the exactly correct answers about half the time) is not surprising. But when combined with either low precision or low recall, we see that the prompt does not perform well. Through an analysis of the generated responses, we found that the model did not appear to have sufficient understanding of the clause and options which resulted in the model trying to force an answer even when the clause did not contain pertinent information. For example, when text discussing change of control only contained a definition of the concept and did not outline any of the obligations of the parties, the model consistently predicted options different from the correct one (i.e., option (g)).
\begin{figure}
 \noindent\fbox{%
    \parbox{\columnwidth}{%
    \fontsize{6}{6}\selectfont
        \textbf{Prompt P1} 
        \hrule 
Read the following [Clause-name] legal clause: 

    [Clause]
     \\
     \\
Pretend you are a party to the agreement in which the [Clause-name] legal clause you have read exists in.
     \\
     \\
You only know what you have read in this prompt.

         [Question]
    \\
    \\
If the clause does not specify, respond with: "Unable to determine". In your response, only include the following most correct groups:
     
         [Options]
\\
\\
In your response, only include the bucket names above. Do not provide an explanation or additional information.
    }%
}  
\caption{Single stage standard prompt \textbf{P1} used for testing the LLM models}
\label{fig:single-prompts} 
\vspace{-1.5em}
\end{figure}

In an attempt to provide more guidance to the model in a structured way, we added a section for specific restrictions that we might like to place on the model. An example of this is prompt \textbf{P2} in  Figure~\ref{fig:single-restrictive-Q1} for the change of control question, \textbf{Q1}. In addition to the clause, the question, and the answer options, it contains a set of instructions that ask the model to be precise, use appropriate legal definitions, and assume a US jurisdiction. These restrictions were motivated by the attempt to help constrain the model's internal processes to those relevant to the question (e.g., different jurisdictions may allow different valid inferences to be made). The prompt also contains more information about changes of control, how certain situations should be interpreted, and special instructions on how option (g) should be treated. The inclusion of the JSON formatting was an attempt to further constrain outputs (e.g., rather than allowing it to generate a free-form list) in addition to improving our ability to parse outputs with fewer edge cases. The prompts for the other questions follow a similar structure and are included in Appendix A.1. 

Unfortunately, these inclusions do not seem to improve the performance of the model (see Table~\ref{tbl:results} for P2). Although this type of restriction helped improve the model's reasoning capabilities in our prior work~\cite{roegiest2023}, it is clear that the questions of interest in this work need more nuanced forms of prompting to elicit better reasoning properties.
\begin{figure}
 \noindent\fbox{%
    \parbox{\columnwidth}{%
    \fontsize{6}{6}\selectfont
        \textbf{Prompt P2}
        \hrule 
You are a legal expert in change of control seeking to answer a specific multiple choice question about change of control.
Your goal is to produce a structured response that will facilitate second-pass review by a human. 
Choosing a wrong answer will waste human effort and should be avoided at all costs. 
\\
\\
\#\#\#\# Question \#\#\#\#

         [Question]
\\
\#\#\#\# Question \#\#\#\#
\\
\\
Which of the following apply:
\\
\\
\#\#\#\# Answer Options \#\#\#\#
     
         [Options]
\\
\#\#\#\# Answer Options \#\#\#\#
\\
\\
\#\#\#\# Change of control text\#\#\#\#

    [Clause]
\\
\#\#\#\# Change of control text \#\#\#\#
\\
\\
\#\#\#\# BEGIN RESTRICTIONS \#\#\#\#
\\
\\
You may select more than one option but you should be certain the answer option is correct if you do so. Prefer being precise over being inclusive.
\\
\\
Use appropriate legal definitions as necessary. You may assume this is in a US jurisdiction. 
\\
\\
Do not use any inferred or implied information, answer based solely on the clause text provided and the answer options. Be as precise as possible when reasoning about which answer options are applicable.
\\
\\
A change of control clause is a clause in an agreement that may require certain actions (including notice, consent, payment of a fee, right to terminate, etc.) of one or more of the parties in the event certain transactions take place during the term of the agreement. These transactions are usually set out in the agreement and may include a sale of all or substantially all of a target company’s assets, any merger of the target company with another company, the transfer of a certain percentage of the target company’s issued and outstanding shares from the target company to the acquirer, in addition to other similar transactions or changes. Certain transactions may be excluded, including those that involve an “affiliate” of the target company. Often, a party will be required to provide notice to or receive the consent of the other party in the event of a change of control. The agreement will likely set out when notice must be provided (either prior to or after the change of control). A party may also be required to receive the consent of the other party prior to entering into a change of control transaction.
\\
\\
Prior to responding, consider that notice of a change of control is distinguishable from notice of termination or notice to terminate. If a party must provide notice of termination, but not notice of change of control, then a party does not have to provide notice of a change of control.
\\
\\
If option (g) is selected, then no other option(s) should be selected.
\\
\\
\#\#\#\# END RESTRICTIONS \#\#\#\#
\\
\\
Return your response as a JSON array with each element taking the following format:
\\
```json
\\
\\
\{
\\
\\
    "bucket" : character  // Respond with the character representing the appropriate bucket.
\\
\}
\\
```
    }%
}  
\caption{Single stage restrictive prompt \textbf{P2} for the change of control question \textbf{Q1}}
\label{fig:single-restrictive-Q1} 
\vspace{-1.5em}
\end{figure}

\subsection{Two Stage Prompts}
Our two stage setup first seeks to produce a summary of the legal text in a way that is relevant to the question with the underlying goal of focusing the actual answer selection on the most relevant attributes of the original clause. Prompt \textbf{P3} (Figure~\ref{fig:twostage-prompts-simple}) illustrates one such prompt for change of control (with other questions being similar). In this summarization stage, we also attempt to avoid hallucinations or other unwarranted inferences or implications by requiring the model to \textit{``only restate what is explicitly provided in the clause and make no inferences or assumptions.''}
An ideal summary is a straightforward and succinct rephrasing of only the question-relevant facts from the legal text without any extraneous information. This should limit potential areas for the model to make assumptions about (seemingly) vaguely worded legal text.
The response from this stage is passed onto a second stage prompt, which is effectively prompt \textbf{P2}. 

\begin{figure}
 \noindent\fbox{%
    \parbox{\columnwidth}{%
    \fontsize{6}{6}\selectfont
        \textbf{Prompt P3}
        \hrule 
\textbf{Stage 1 }
\\
You are a legal expert specializing in change of control.
\\
Given the following change of control text, provide a simple but thorough response for the provided question.

You should provide enough detail to allow subsequent decisions to be made from the response but only restate what is explicitly provided in the clause and make no inferences or assumptions. 
\\
\\
\#\#\#\# BEGIN QUESTION \#\#\#\#

         [Question]
\\
\#\#\#\# END QUESTION \#\#\#\#
\\
\\
\#\#\#\# BEGIN TEXT\#\#\#\#

    [Clause]
\\
\#\#\#\# END TEXT \#\#\#\#

        \hrule 
 
\textbf{Stage 2}
\\
You are a legal expert seeking to answer a specific multiple choice question about change of control.

The following is an answer to the question: "[Question]". 
\\
\\
\#\#\#\# BEGIN RESPONSE\#\#\#\#

    [Response]
\\
\#\#\#\# END RESPONSE \#\#\#\#
\\
\\
Based upon this response, which of the following apply:
\\
\#\#\#\# Answer Options \#\#\#\#
     
         [Options]
\\
\#\#\#\# Answer Options \#\#\#\#
\\
\\
\#\#\#\# BEGIN RESTRICTIONS \#\#\#\#
\\
\\
You may select more than one option but you should be certain the answer option is correct if you do so. Prefer being precise over being inclusive.
\\
\\
Use appropriate legal definitions as necessary. You may assume this is in a US jurisdiction.  
\\
\\
Do not use any inferred or implied information, answer based solely on the clause text provided and the answer options. Be as precise as possible when reasoning about which answer options are applicable.
\\
\\
A change of control clause is a clause in an agreement that may require certain actions (including notice, consent, payment of a fee, right to terminate, etc.) of one or more of the parties in the event certain transactions take place during the term of the agreement. These transactions are usually set out in the agreement and may include a sale of all or substantially all of a target company’s assets, any merger of the target company with another company, the transfer of a certain percentage of the target company’s issued and outstanding shares from the target company to the acquirer, in addition to other similar transactions or changes. Certain transactions may be excluded, including those that involve an “affiliate” of the target company.
\\
\\
When one party is required to give notice to the other party, they must send the other party a letter/email setting out what change will or has taken place. Consent requires one party to notify the other party of the proposed change and actually receive the other party’s consent prior to pursuing the change.
\\
\\
Prior to responding, consider that notice of a change of control is distinguishable from notice of termination or notice to terminate. If a party must provide notice of termination, but not notice of change of control, then a party does not have to provide notice of a change of control.
\\
\\
If option (g) is selected, then no other option(s) should be selected.
\\
\\
\#\#\#\# END RESTRICTIONS \#\#\#\#
\\
\\
Return your response as a JSON array with each element taking the following format:
\\
```json
\\
\\
\{
\\
\\
    "bucket" : character  // Respond with the character representing the appropriate bucket.
\\
\}
\\
```
    }%
}  
\caption{Simple two stage prompt \textbf{P3} for the change of control question \textbf{Q1}}
\label{fig:twostage-prompts-simple} 
\vspace{-1.5em}
\end{figure}

From these changes, we see a minor improvement in precision and recall, especially for the questions \textbf{Q1} and \textbf{Q2}. While not as dramatic as one would like, in examining the generated summaries, we found that the model was able to better direct the model in the second stage to focus only on relevant data points and removed areas for interpretation as desired. Figure~\ref{fig:responses} shows an example change of control clause, extracted from an agreement between two parties FoundryCo and AMD\footnote{\url{https://ir.amd.com/sec-filings/content/0000002488-20-000164/exh101amdq310q20.htm}}. The clause first provides the definition of change of control in the context of the agreement. It then states that AMD is not permitted to engage with a third party (name redacted in the published agreement) as a second source manufacturer of any MPU Products, subject to a reasonable wind-down period, when there is a change of control. This is an obligation other than providing notice or getting consent, therefore the expected answer is option (c).  The summary generated by the GPT-4o model lists the definition and the obligations upon change of control. However, it gets the answer only partially correct, because it assumes that the agreement remains in effect although the clause does not explicitly say so.\footnote{We note that this is particularly nuanced and caused internal debate about when this type of answer is appropriate before settling on the current version.} 
\begin{figure}
 \noindent\fbox{%
    \parbox{\columnwidth}{%
    \fontsize{6}{6}\selectfont
        \textbf{Prompt P4}
        \hrule 
\textbf{Stage 1 }
\\
You are a legal expert specializing in change of control.
\\
Given the following change of control text, provide a simple but thorough response for the provided question.

You should provide enough detail to allow subsequent decisions to be made from the response but only restate what is explicitly provided in the clause and make no inferences or assumptions. 
\\
\\
\#\#\#\# BEGIN QUESTION \#\#\#\#

         [Question]
\\
\#\#\#\# END QUESTION \#\#\#\#
\\
\\
\#\#\#\# BEGIN TEXT\#\#\#\#

    [Clause]
\\
\#\#\#\# END TEXT \#\#\#\#
\\
\\
\#\#\#\# BEGIN RESTRICTIONS \#\#\#\#
\\
A change of control clause is a clause in an agreement that may require certain actions (including notice, consent, payment of a fee, right to terminate, etc.) of one or more of the parties in the event certain transactions take place during the term of the agreement. These transactions are usually set out in the agreement and may include a sale of all or substantially all of a target company’s assets, any merger of the target company with another company, the transfer of a certain percentage of the target company’s issued and outstanding shares from the target company to the acquirer, in addition to other similar transactions or changes. Certain transactions may be excluded, including those that involve an “affiliate” of the target company.
\\
\\
When one party is required to give notice to the other party, they must send the other party a letter/email setting out what change will or has taken place. Consent requires one party to notify the other party of the proposed change and actually receive the other party’s consent prior to pursuing the change.
\\
\\
Prior to responding, consider that notice of a change of control is distinguishable from notice of termination or notice to terminate. If a party must provide notice of termination, but not notice of change of control, then a party does not have to provide notice of a change of control.
\\
\\
\#\#\#\# END RESTRICTIONS \#\#\#\#
\\
\\
Your response will be subsequently mapped to the following options, so be sure that your answer is thorough:
     
         [Options]

    }%
}  
\caption{The first stage of a more restrictive two stage prompt \textbf{P4} for the change of control question \textbf{Q1}. The second stage is identical to that of prompt \textbf{P3}.}
\label{fig:twostage-prompts-complex} 
\vspace{-1.5em}
\end{figure}

We subsequently modified the first stage by adding in similar restrictions to the single stage prompt but also included the answer options as a reference for the model to use when crafting a summary. In this way, we sought to help guide the model to even more carefully tailor the summary to align with the subsequent answer selection. Prompt \textbf{P4} (Figure~\ref{fig:twostage-prompts-complex}) illustrates these changes with the second stage being identical to that of \textbf{P3}. As expected, the inclusion of the answer options guided the model to produce more specific summaries with the model often making mention of which options would be appropriate. However, we find it interesting that the format of the summaries varied non-trivially from one piece of legal text to the next (e.g., several paragraphs, bullet point lists, lists of applicable answers followed by an explanation of why each is applicable) rather than having a single stable format. Regardless, this inclusion generally improved effectiveness across the board with the models producing the appropriate answers more often. For questions \textbf{Q1} and \textbf{Q3}, there is a substantial increase in the exact match accuracy. For question \textbf{Q2}, although there is no substantial change in the exact match accuracy, the recall improves by 15\% for model GPT-4o and 5\% for model GPT-4 Turbo. 

In the example shown in Figure~\ref{fig:responses}, we can see that the summary generated by the GPT-4o model for prompt \textbf{P4} is quite different from that generated for prompt \textbf{P3}. The \textbf{P4} summary is focused on the obligations of the parties post change of control and does not include the definition. It includes a reference to the chosen option, and provides an ``explanation'' of why that option was chosen. This summary is similar to the explanation generated by the model when it is asked directly why a particular option was chosen. The results show that providing more context to the model during both summarization and option selection can be helpful in generating an accurate and tailored summary, and subsequently in choosing the right option. 

\section{Discussion}
\subsection{Limitations of Two Stages}
Across all formulations, the two stage prompting strategy does not improve any of the metrics for question \textbf{Q4}. We investigated this further by looking into which answer options were not being correctly predicted. Table~\ref{tbl:force_majeure} shows the precision and recall for each of the answer options. We found that both GPT-4o and GPT-4 Turbo models consistently perform poorly while predicting option (d), thus, reducing average precision and recall. We examined the clauses for which this answer option was applicable and found significant linguistic variation in these clauses. These variations correspond to the fact that ``utility failures'' can refer to many different things and may rely, at least partially, on the subjective view of the annotator/question builder rather than a single objective definition (i.e., a legal interpretation of ``utility'' may not correspond to a colloquial one that a generative model saw frequently in training data). 


    We included an exhaustive definition of utility failures, which lists out all possible interpretations of utility failures we found in the clauses, into prompt \textbf{P4} (Figure 10 in Appendix A.1) and found that the exact match accuracy is slightly improved to 70.4\% and 74.8\% for GPT-4o and GPT-4-Turbo, respectively. This is an increase of 2.6\% and 1.7\% over the original prompt \textbf{P4}. 
GPT-4-Turbo  yields precision and recall both equal to 0.96, which is a reduction of 1 point in precision and an increase of 4 points in recall over the 
original prompt \textbf{P4}. In Appendix A.2, we show an example of how the inclusion of the exhaustive definition changes the generated summary and thereby leads to a correct answer. 

While there are some improvements, this type of prompt is far from easy to build, and may not generalize well because it relies on including all possible language variations in the clauses. Indeed, this approach mimics a ``rules-like'' approach and we argue that applying this prompt to further examples of force majeure with utility failures may well reveal further variations that need to be captured (i.e., similar to rules-based approaches). We conclude that two stage prompts are unable to perform well when answering must rely on understanding a large amount of linguistic variation. 

\begin{table}[ht]
\scriptsize
  \centering
\resizebox{\columnwidth}{!}{
  \begin{tabular}{|c||c||c|c||c|c||c|c||c|c|}
  \hline
  \textbf{Model}  & \textbf{Option}  & \multicolumn{2}{c||}{\textbf{P1}} & \multicolumn{2}{c||}{\textbf{P2}} & \multicolumn{2}{c||}{\textbf{P3}}  & \multicolumn{2}{c|}{\textbf{P4}} \\
  \hline
  & & P & R& P & R& P & R& P & R\\
  \hline
\textbf{GPT-4o} & A &   1.00&  1.00 &  1.00&0.95  &1.00 & 0.97 & 1.00&  0.96\\
\cline{2-10}
 & B & 1.00  &0.99  & 1.00 &0.96  &1.00&  0.97& 1.00&  0.96  \\
 \cline{2-10}
 & C &  1.00 & 0.96 & 1.00 & 1.00 &1.00 & 1.00 &1.00 & 0.88 \\
 \cline{2-10}
 & D & 0.76 & 0.60 & 1.00 & 0.44 & 0.79&0.79  & 0.77&  0.84\\
 \cline{2-10}
 & E & 0.98 & 0.92  &0.98  & 0.85 &0.96 & 0.95 &0.97 & 0.95 \\
 \cline{2-10}
 & F & 0.97 & 0.98 &0.98  & 0.98 &0.99 & 0.67 & 0.96&1.00  \\
 \hline
 \textbf{GPT-4} & A & 1.00  &  1.00 & 1.00 & 1.00 &1.00   &1.00  & 1.00& 0.98 \\
\cline{2-10}
\textbf{Turbo} & B &  1.00 & 0.99 & 1.00 & 0.96 &1.00 & 0.99 &1.00 &  0.97\\
 \cline{2-10}
 & C & 1.00  & 1.00  & 1.00 & 1.00 &1.00 &  1.00&1.00 & 0.96 \\
 \cline{2-10}
 & D & 0.82 &0.77  & 0.83 & 0.67 &0.83 & 0.67 &0.86 &0.72  \\
 \cline{2-10}
 & E & 0.97 & 0.89 & 0.98 &  0.89&0.97 & 0.89 &0.97 & 0.89 \\
 \cline{2-10}
 & F & 0.95 &  0.97 & 0.97 &1.00  & 0.99&  0.76& 0.97&0.97  \\
  
    \hline
  \end{tabular}}
  \caption{Precision (P) and recall (R) of the LLM models GPT-4o and GPT-4 Turbo on different answer options of the force majeure question \textbf{Q4} using the four prompts.}
  \label{tbl:force_majeure}
  \vspace{-2em}
\end{table}

\begin{figure*} 
    \centering
    \includegraphics[width=\textwidth]{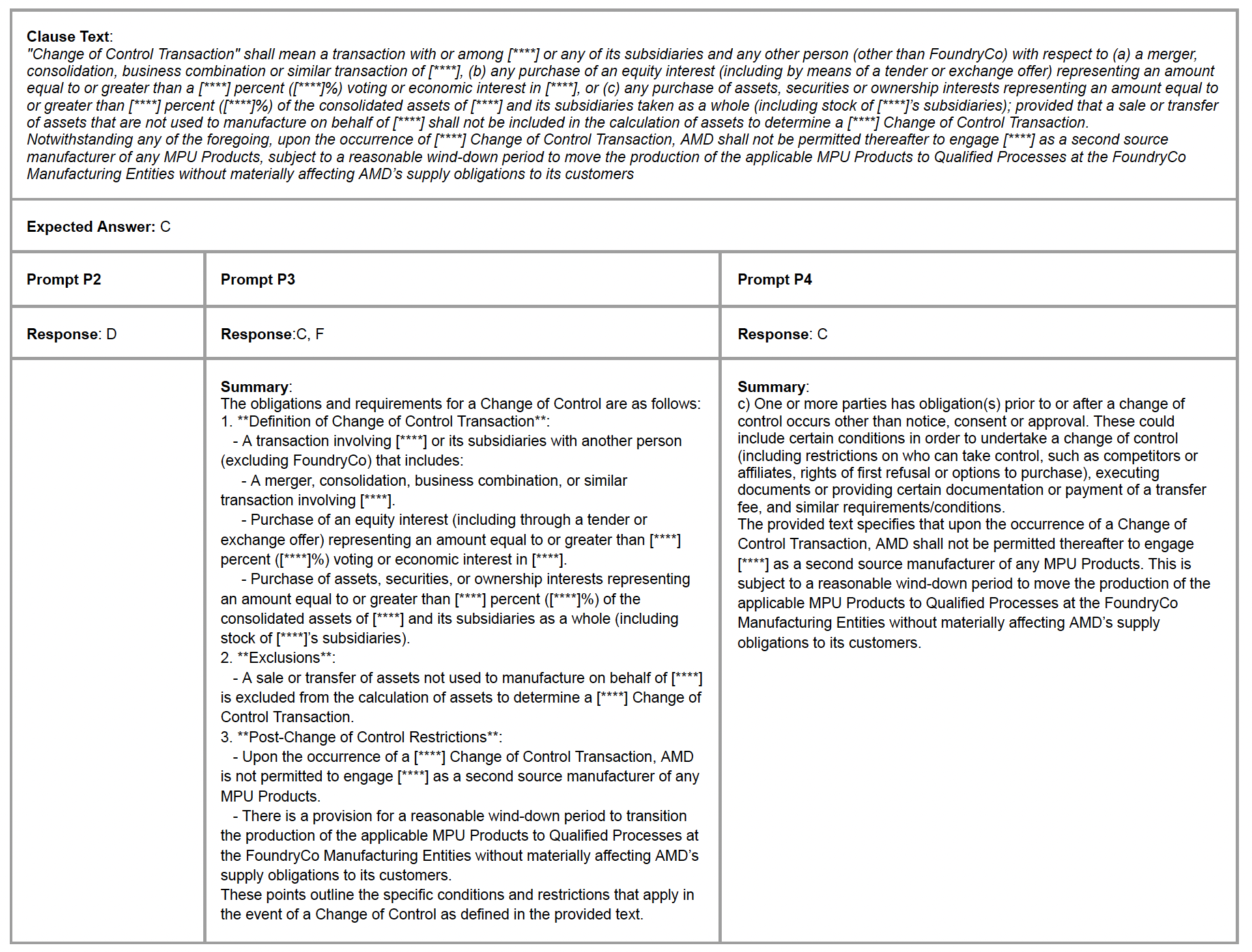}
    \caption{Responses of the GPT-4o model on a sample change of control clause for each prompt. The single-stage prompt gets the answer wrong. The simple two-stage prompt \textbf{P3} gets the answer partially correct due to the generated precise summary of the clause. Prompt \textbf{P4} further aligns the summary toward the answer options, leading to a correct answer choice.  }
    \label{fig:responses}
    \vspace{-1.5em}
\end{figure*}

\subsection{On Summary Variations}
We remarked earlier in this work that the summaries produced by \textbf{P3} and \textbf{P4} tended to be fairly varied in format with no particularly consistent trends. In analyzing this more thoroughly, we find that the type of question plays an important factor in what styles of summaries get generated. For reasoning-style questions (e.g., change of control, assignment) there is a strong trend for summaries (75\% of them) to be longer forms of exposition (i.e., explanations, long form summaries) but for more ``list-like'' questions (e.g., force majeure), there is a corresponding bias for list-like responses. While we do not find substantive gains from the variations in summaries with respect to subsequent answer selection, more investigation is warranted to see if there are ways to leverage these different styles of summaries more effectively.

\subsection{Beyond Summarization}
In our two stage prompts, we first utilize summarization as the main technique to help narrow the focus of the subsequent stage. As we have seen, this has generally worked well for these seemingly hard to answer questions. 
However, it is possible that other forms of re-framing the legal text may work. For example, we initially thought that explicitly specifying rights and obligations on a per-party basis for change of control held promise. Essentially, if all these details are clearly spelled out then it should be easier for the model to succeed at mapping them to answers. 

We tested this idea with our change of control question and rather than asking for a summary, explicitly asked it to ``Provide for each party, its requirements, obligations, and duties when a change of control occurs.'' While the effects are not necessarily complete failure (e.g., GPT-4-Turbo results are 16 points lower on exact match accuracy, 10 points lower on Precision, and 2 points higher on Recall), we find in examining the results that this prompt essentially allows the model to infer many different obligations than are explicitly stated (e.g., incorrect option (f) supporting evidence is frequently generated). This is not to say that such a prompt could not work but may then need specific tailoring of options to the results produced. For example, with this prompt, it may be easier to provide a party and then ask which of the options applies to that party rather than trying to provide a general overview of the obligations when a change of control occurs. As our needs continue to develop, we plan to explore this direction more fully. 

Along similar lines, we noticed that for prompt \textbf{P4}, the summaries contained explanation-like summaries of why certain options should be chosen. This led us to wonder whether a single stage prompt that encouraged models to produce a structured explanation per answer option would yield similar improvement. That is, if we required a justification per answer would this result in similar increases in effectiveness to the highly tailored summary? In conducting a small test with \textbf{Q1}, in which we modified \textbf{P1} with the extra explanation field in the JSON output,  we found that the generated explanations aligned fairly closely with the summaries produced by \textbf{P4} but that effectiveness dropped overall. For GPT-4o, exact accuracy dropped 15 points, average precision dropped 9 points, and average recall increasing 6 points. The increase of recall is not entirely surprising as \textbf{P1} historically had the best recall owing to its generally unrestricted nature. The drop in exact match accuracy and precision likely correlates to the increase in recall, where the model is simply being more liberal in its answer selection. We suspect, but plan to investigate further, that this is due to the generative nature of these models. 

\section{Conclusion and Future Work}

We have presented a study of legal question answering with a focus on hard-to-answer legal questions. We presented an in-depth analysis of different prompting styles and an exploration of variations that enabled more accurate answers to questions while attempting to minimize costs. We have found that for questions about legal text filled with a mix of relevant and non-relevant content, two-stage prompts that produce summaries tailored to the question and desired structured outputs can help LLMs focus on desired attributes and minimize inference of unspecified details. We have also seen that such prompts may not work when some answers require highly specialized and detailed responses due to the linguistic variations present in the underlying text. However, using one and two-stage prompts can help identify when such cases occur as no improvement would be seen for such cases. Finally, we explore some additional variations to our two-stage strategy that suggest avenues for future investigation into the better techniques to use when targeting different use cases. 

In the future, we also plan to investigate if it is possible to build templates using the definition of the underlying clauses and placing some standard restrictions on the model. We also plan to explore if pre-training or fine-tuning the LLM model with contracts and legal questions and answers would help the LLM understand the nuances in legal language and produce more relevant summaries, thereby improving the performance of the model. Finally, we plan to
investigate the robustness of our prompts to different versions of generative models.

\newpage
\newpage
\bibliography{main}
\newpage
\appendix
\section{Appendix}
\subsection{Prompts for questions \textbf{Q2}, \textbf{Q3} and \textbf{Q4}}
\label{sec:appendix-prompts}
Figures~\ref{fig:single-prompt-Q2},~\ref{fig:single-prompt-Q3}, and~\ref{fig:single-prompt-Q4} show the restrictive single stage prompt \textbf{P2} used for questions \textbf{Q2}, \textbf{Q3} and \textbf{Q4}.

Figures~\ref{fig:twostage-prompts-simple-Q2},~\ref{fig:twostage-prompts-simple-Q3}, and~\ref{fig:twostage-prompts-simple-Q4} show the simple two stage prompt \textbf{P3} used for questions \textbf{Q2}, \textbf{Q3} and \textbf{Q4}.

Figures~\ref{fig:twostage-prompts-complex-Q2},~\ref{fig:twostage-prompts-complex-Q3}, and~\ref{fig:twostage-prompts-complex-Q4} show the first stage of the restrictive two stage prompt \textbf{P4} used for questions \textbf{Q2}, \textbf{Q3} and \textbf{Q4}. The second stage of \textbf{P4} is the same as that in prompt \textbf{P3}.

Figure~\ref{fig:twostage-prompts-complex-force-majeure} shows the two stage prompt \textbf{P4} modified to include an exhaustive definition of utility failures to improve the effectiveness of the LLM models on question \textbf{Q4}. 

\subsection{Analysis of the prompt effectiveness on the Force Majeure question}
\label{sec:appendix-force-majeure}
In Figure~\ref{fig:force-majeure-analysis}, we show a sample Force Majeure clause which includes multiple triggers including "\textit{fire or other casualty or accident, strikes or labor disputes, problems in procurement of raw materials, power or supplies, war or other violence, any law, order, proclamation, regulation, ordinance, demand or requirement of governmental agency or intergovernmental body, or any other act or condition whatsoever beyond the reasonable control of a party}". The expected answer options are (a), (b), (d), (e) and (f). When using prompt \textbf{P4}, the model is unable to predict options (a) and (d). On the other hand, when using prompt \textbf{P5} which includes a list of all events that are considered utility failures, the model is able to map "\textit{problems in procurement of raw materials}"  to utility failures, and predict the expected answer.

\newpage
\begin{figure*}
 \noindent\fbox{%
    \parbox{\textwidth}{%
    \fontsize{8}{8}\selectfont
        \textbf{Prompt P2}
        \hrule 
\vspace{0.5em}
You are a legal expert specializing in assignment. 
Given the following assignment clause,  which of the answer options apply as the answer to the question "[Question]":
\\
\\
\#\#\#\# BEGIN TEXT\#\#\#\#

    [Clause]
\\
\#\#\#\# END TEXT \#\#\#\#
\\
\\
\#\#\#\# Answer Options \#\#\#\#
     
         [Options]
\\
\#\#\#\# Answer Options \#\#\#\#
\\
\\
\#\#\#\# BEGIN RESTRICTIONS \#\#\#\#
\\
\\
You may select more than one option but you should be certain the answer option is correct if you do so. Prefer being precise over being inclusive.
\\
\\
Use appropriate legal definitions as necessary. You may assume this is in a US jurisdiction. 
\\
\\
Do not use any inferred or implied information, answer based solely on the clause text provided and the answer options. Be as precise as possible when reasoning about which answer options are applicable.
\\
\\
An assignment clause is a clause in an agreement that regulates the extent to which a party's interest in an agreement may be assigned to another party. The clause may require certain actions (including notice, consent, payment of a fee, right to terminate, etc.) of one or more of the parties in the event certain transactions take place during the term of the agreement.
\\
\\
When one party is required to give notice to the other party, they must send the other party a letter/email setting out what change will or has taken place. Consent requires one party to notify the other party of the proposed change and actually receive the other party’s consent prior to pursuing the change.
\\
\\
Prior to responding, consider that notice of an assignment is distinguishable from notice of termination or notice to terminate. If a party must provide notice of termination, but not notice of assignment, then a party does not have to provide notice of an assignment.
\\
\\
Also note that if the agreement terminates upon assignment, the termination may be automatic (i.e. the agreement becomes void upon assignment) or initiated by either parties. Option (d) should be selected in either of these cases. 
\\
\\
If the clause does not explicitly say if or how the agreement can be assigned, respond with: "Unable to determine".
\\
\\
\#\#\#\# END RESTRICTIONS \#\#\#\#
\\
\\
Return your response as a JSON array with each element taking the following format:
\\
```json
\\
\\
\{
\\
\\
    "bucket" : character  // Respond with the character representing the appropriate bucket.
\\
\}
\\
```
    }%
}  
\caption{Restrictive single stage prompt \textbf{P2} for question \textbf{Q2} "What are the obligations of the parties for assigning the agreement?"}
\label{fig:single-prompt-Q2} 
\vspace{-1.5em}
\end{figure*}

\begin{figure*}
 \noindent\fbox{%
    \parbox{\textwidth}{%
    \fontsize{8}{8}\selectfont
        \textbf{Prompt P2}
        \hrule 
\vspace{0.5em}
You are a legal expert specializing in insurance. 
Given the following insurance clause,  which of the answer options apply as the answer to the question "[Question]":
\\
\\
\#\#\#\# BEGIN TEXT\#\#\#\#

    [Clause]
\\
\#\#\#\# END TEXT \#\#\#\#
\\
\\
\#\#\#\# Answer Options \#\#\#\#
     
         [Options]
\\
\#\#\#\# Answer Options \#\#\#\#
\\
\\
\#\#\#\# BEGIN RESTRICTIONS \#\#\#\#
\\
\\
You may select more than one option but you should be certain the answer option is correct if you do so. Prefer being precise over being inclusive.
\\
\\
Use appropriate legal definitions as necessary. You may assume this is in a US jurisdiction. 
\\
\\
Do not use any inferred or implied information, answer based solely on the clause text provided and the answer options. Be as precise as possible when reasoning about which answer options are applicable.
\\
\\
If the clause contains a type of insurance which is not included in the above options, include "k) Other type(s) of insurance that are not listed in the above options." in the response.  
\\
\\
If the clause does not explicitly state what type of insurance(s) a party must have, respond with: "Unable to determine".
\\
\\
\#\#\#\# END RESTRICTIONS \#\#\#\#
\\
\\
Return your response as a JSON array with each element taking the following format:
\\
```json
\\
\\
\{
\\
\\
    "bucket" : character  // Respond with the character representing the appropriate bucket.
\\
\}
\\
```
    }%
}  
\caption{Restrictive single stage prompt \textbf{P2} for question \textbf{Q3} "What type(s) of insurance is a party required to have?"}
\label{fig:single-prompt-Q3} 
\vspace{-1.5em}
\end{figure*}
\begin{figure*}
 \noindent\fbox{%
    \parbox{\textwidth}{%
    \fontsize{8}{8}\selectfont
        \textbf{Prompt P2}
        \hrule 
\vspace{0.5em}
You are a legal expert specializing in Force Majeure. 
Given the following Force Majeure clause,  which of the answer options apply as the answer to the question "[Question]":
\\
\\
\#\#\#\# BEGIN TEXT\#\#\#\#

    [Clause]
\\
\#\#\#\# END TEXT \#\#\#\#
\\
\\
\#\#\#\# Answer Options \#\#\#\#
     
         [Options]
\\
\#\#\#\# Answer Options \#\#\#\#
\\
\\
\#\#\#\# BEGIN RESTRICTIONS \#\#\#\#
\\
\\
You may select more than one option but you should be certain the answer option is correct if you do so. Prefer being precise over being inclusive.
\\
\\
Use appropriate legal definitions as necessary. You may assume this is in a US jurisdiction. 
\\
\\
Do not use any inferred or implied information, answer based solely on the clause text provided and the answer options. Be as precise as possible when reasoning about which answer options are applicable.
\\
\\
If the clause contains catch-all language for all types of 
events beyond the reasonable control of the parties, include "f) There is catch-all language in the clause" in the response.   
\\
\\
If the clause does not explicitly state what events are included in the force majeure events, respond with: "Unable to determine".
\\
\\
\#\#\#\# END RESTRICTIONS \#\#\#\#
\\
\\
Return your response as a JSON array with each element taking the following format:
\\
```json
\\
\\
\{
\\
\\
    "bucket" : character  // Respond with the character representing the appropriate bucket.
\\
\}
\\
```
    }%
}  
\caption{Restrictive single stage prompt \textbf{P2} for question \textbf{Q4} "What are the force majeure trigger events?"}
\label{fig:single-prompt-Q4} 
\vspace{-1.5em}
\end{figure*}
\begin{figure*}
 \noindent\fbox{%
    \parbox{\textwidth}{%
    \fontsize{8}{8}\selectfont
        \textbf{Prompt P3}
        \hrule 
\vspace{1em}
\textbf{Stage 1 }
\\
You are a legal expert specializing in assignment.
\\
Given the following assignment text, provide a simple but thorough response for the provided question.

You should provide enough detail to allow subsequent decisions to be made from the response but only restate what is explicitly provided in the clause and make no inferences or assumptions. 
\\
\\
\#\#\#\# BEGIN QUESTION \#\#\#\#

         [Question]
\\
\#\#\#\# END QUESTION \#\#\#\#
\\
\\
\#\#\#\# BEGIN TEXT\#\#\#\#

    [Clause]
\\
\#\#\#\# END TEXT \#\#\#\#
\hrule 
\vspace{1em}
\textbf{Stage 2}
\\
You are a legal expert seeking to answer a specific multiple choice question about assignment.

The following is an answer to the question: "[Question]". 
\\
\\
\#\#\#\# BEGIN RESPONSE\#\#\#\#

    [Response]
\\
\#\#\#\# END RESPONSE \#\#\#\#
\\
\\
Based upon this response, which of the following apply:
\\
\#\#\#\# Answer Options \#\#\#\#
     
         [Options]
\\
\#\#\#\# Answer Options \#\#\#\#
\\
\\
\#\#\#\# BEGIN RESTRICTIONS \#\#\#\#
\\
\\
You may select more than one option but you should be certain the answer option is correct if you do so. Prefer being precise over being inclusive.
\\
\\
Use appropriate legal definitions as necessary. You may assume this is in a US jurisdiction.  
\\
\\
Do not use any inferred or implied information, answer based solely on the clause text provided and the answer options. Be as precise as possible when reasoning about which answer options are applicable.
\\
\\
An assignment clause is a clause in an agreement that regulates the extent to which a party's interest in an agreement may be assigned to another party. The clause may require certain actions (including notice, consent, payment of a fee, right to terminate, etc.) of one or more of the parties in the event certain transactions take place during the term of the agreement.
\\
\\
When one party is required to give notice to the other party, they must send the other party a letter/email setting out what change will or has taken place. Consent requires one party to notify the other party of the proposed change and actually receive the other party’s consent prior to pursuing the change.
\\
\\
Prior to responding, consider that notice of an assignment is distinguishable from notice of termination or notice to terminate. If a party must provide notice of termination, but not notice of assignment, then a party does not have to provide notice of an assignment.
\\
\\
Also note that if the agreement terminates upon assignment, the termination may be automatic (i.e. the agreement becomes void upon assignment) or initiated by either parties. Option (d) should be selected in either of these cases. 
\\
\\
\#\#\#\# END RESTRICTIONS \#\#\#\#
\\
\\
Return your response as a JSON array with each element taking the following format:
\\
```json
\\
\\
\{
\\
\\
    "bucket" : character  // Respond with the character representing the appropriate bucket.
\\
\}
\\
```
    }%
}  
\caption{Simple two stage prompt \textbf{P3} for question \textbf{Q2} "What are the obligations of the parties for assigning the agreement?"}
\label{fig:twostage-prompts-simple-Q2} 
\vspace{-1.5em}
\end{figure*}

\begin{figure*}
 \noindent\fbox{%
    \parbox{\textwidth}{%
    \fontsize{8}{8}\selectfont
        \textbf{Prompt P3}
        \hrule 
\vspace{1em}
\textbf{Stage 1 }
\\
You are a legal expert specializing in insurance.
\\
Given the following insurance clause, provide a simple but thorough response for the provided question.

You should provide enough detail to allow subsequent decisions to be made from the response but only restate what is explicitly provided in the clause and make no inferences or assumptions. 
\\
\\
\#\#\#\# BEGIN QUESTION \#\#\#\#

         [Question]
\\
\#\#\#\# END QUESTION \#\#\#\#
\\
\\
\#\#\#\# BEGIN TEXT\#\#\#\#

    [Clause]
\\
\#\#\#\# END TEXT \#\#\#\#
\hrule
\vspace{1em}
\textbf{Stage 2}
\\
You are a legal expert seeking to answer a specific multiple choice question about insurance.

The following is an answer to the question: "[Question]". 
\\
\\
\#\#\#\# BEGIN RESPONSE\#\#\#\#

    [Response]
\\
\#\#\#\# END RESPONSE \#\#\#\#
\\
\\
Based upon this response, which of the following apply as the answer to the question "[Question]":
\\
\#\#\#\# Answer Options \#\#\#\#
     
         [Options]
\\
\#\#\#\# Answer Options \#\#\#\#
\\
\\
\#\#\#\# BEGIN RESTRICTIONS \#\#\#\#
\\
\\
You may select more than one option but you should be certain the answer option is correct if you do so. Prefer being precise over being inclusive.
\\
\\
Use appropriate legal definitions as necessary. You may assume this is in a US jurisdiction.  
\\
\\
Do not use any inferred or implied information, answer based solely on the clause text provided and the answer options. Be as precise as possible when reasoning about which answer options are applicable.
\\
\\
If the clause contains a type of insurance which is not included in the above options as part of the list you created, include "i) Other type(s) of insurance that are not listed in the above options." in the response. If the clause does not explicitly state what type(s) of insurance a party must have, respond with: "Unable to determine".
\\
\\
\#\#\#\# END RESTRICTIONS \#\#\#\#
\\
\\
Return your response as a JSON array with each element taking the following format:
\\
```json
\\
\\
\{
\\
\\
    "bucket" : character  // Respond with the character representing the appropriate bucket.
\\
\}
\\
```
    }%
}  
\caption{Simple two stage prompt \textbf{P3} for question \textbf{Q3} "What type(s) of insurance is a party required to have?"}
\label{fig:twostage-prompts-simple-Q3} 
\vspace{-1.5em}
\end{figure*}
\begin{figure*}
 \noindent\fbox{%
    \parbox{\textwidth}{%
    \fontsize{8}{8}\selectfont
        \textbf{Prompt P3}
        \hrule 
\vspace{1em}
\textbf{Stage 1 }
\\
You are a legal expert specializing in Force Majeure.
\\
Given the following Force Majeure clause, provide a simple but thorough response for the provided question.

You should provide enough detail to allow subsequent decisions to be made from the response but only restate what is explicitly provided in the clause and make no inferences or assumptions. 
\\
\\
\#\#\#\# BEGIN QUESTION \#\#\#\#

         [Question]
\\
\#\#\#\# END QUESTION \#\#\#\#
\\
\\
\#\#\#\# BEGIN TEXT\#\#\#\#

    [Clause]
\\
\#\#\#\# END TEXT \#\#\#\#

        \hrule 

\vspace{1em}
\textbf{Stage 2}
\\
You are a legal expert seeking to answer a specific multiple choice question about Force Majeure.

The following is an answer to the question: "[Question]". 
\\
\\
\#\#\#\# BEGIN RESPONSE\#\#\#\#

    [Response]
\\
\#\#\#\# END RESPONSE \#\#\#\#
\\
\\
Based upon this response, which of the following apply as the answer to the question "[Question]":
\\
\#\#\#\# Answer Options \#\#\#\#
     
         [Options]
\\
\#\#\#\# Answer Options \#\#\#\#
\\
\\
\#\#\#\# BEGIN RESTRICTIONS \#\#\#\#
\\
\\
You may select more than one option but you should be certain the answer option is correct if you do so. Prefer being precise over being inclusive.
\\
\\
Use appropriate legal definitions as necessary. You may assume this is in a US jurisdiction.  
\\
\\
Do not use any inferred or implied information, answer based solely on the clause text provided and the answer options. Be as precise as possible when reasoning about which answer options are applicable.
\\
\\
A force majeure clause in an agreement specifies the events and circumstances (such as acts of God, war, utility failures, public health events like epidemics and pandemics, labour disruptions, etc.) beyond the reasonable control of the parties for which the parties are not liable.
\\
\\
When responding, note that the "utilities" is defined as: energy, power, fuel, internet, communications, network, and any other utilities. 
"Failures" is defined as: any shortages, failures, lack of, inability to obtain, restrictions, outages, blackouts, or similar. 
Include option (d) in your response only for utility failures as defined above.
\\
\\
If the response states that there is catch-all language in the clause, include "f) There is catch-all language in the clause" in the response. 
If the clause does not explicitly state what events are included in the force majeure events, respond with: "Unable to determine".
\\
\\
\#\#\#\# END RESTRICTIONS \#\#\#\#
\\
\\
Return your response as a JSON array with each element taking the following format:
\\
```json
\\
\\
\{
\\
\\
    "bucket" : character  // Respond with the character representing the appropriate bucket.
\\
\}
\\
```
    }%
}  
\caption{Simple two stage prompt \textbf{P3} for question \textbf{Q4} "What are the force majeure trigger events?"}
\label{fig:twostage-prompts-simple-Q4} 
\vspace{-1.5em}
\end{figure*}
\begin{figure*}
 \noindent\fbox{%
    \parbox{\textwidth}{%
    \fontsize{8}{8}\selectfont
        \textbf{Prompt P4}
        \hrule 
\vspace{1em}
\textbf{Stage 1 }
\\
You are a legal expert specializing in assignment.
\\
Given the following assignment text, provide a simple but thorough response for the provided question.

You should provide enough detail to allow subsequent decisions to be made from the response but only restate what is explicitly provided in the clause and make no inferences or assumptions. 
\\
\\
\#\#\#\# BEGIN QUESTION \#\#\#\#

         [Question]
\\
\#\#\#\# END QUESTION \#\#\#\#
\\
\\
\#\#\#\# BEGIN TEXT\#\#\#\#

    [Clause]
\\
\#\#\#\# END TEXT \#\#\#\#
\\
\\
\#\#\#\# BEGIN RESTRICTIONS \#\#\#\#
\\
You may select more than one option but you should be certain the answer option is correct if you do so. Prefer being precise over being inclusive.
\\
\\
Use appropriate legal definitions as necessary. You may assume this is in a US jurisdiction.
\\
\\
Do not use any inferred or implied information, answer based solely on the clause text provided and the answer options. Be as precise as possible when reasoning about which answer options are applicable.
\\
\\
An assignment clause is a clause in an agreement that regulates the extent to which a party's interest in an agreement may be assigned to another party. The clause may require certain actions (including notice, consent, payment of a fee, right to terminate, etc.) of one or more of the parties in the event certain transactions take place during the term of the agreement. 
\\
\\
When one party is required to give notice to the other party, they must send the other party a letter/email setting out what change will or has taken place. Consent requires one party to notify the other party of the proposed change and actually receive the other party’s consent prior to pursuing the assignment.
\\
\\
Prior to responding, consider that notice of an assignment is distinguishable from notice of termination or notice to terminate. If a party must provide notice of termination, but not notice of assignment, then a party does not have to provide notice of an assignment.
\\
\\
Also note that if the agreement terminates upon assignment, the termination may be automatic (i.e. the agreement becomes void upon assignment) or initiated by either parties. 
\#\#\#\# END RESTRICTIONS \#\#\#\#
\\
\\
Your response will be subsequently mapped to the following options, so be sure that your answer is thorough:
     
         [Options]

    }%
}  
\caption{The first stage of a more restrictive two stage prompt \textbf{P4} for question \textbf{Q2} "What are the obligations of the parties for assigning the agreement?". The second stage is identical to that of prompt \textbf{P3}.}
\label{fig:twostage-prompts-complex-Q2} 
\vspace{-1.5em}
\end{figure*}

\begin{figure*}
 \noindent\fbox{%
    \parbox{\textwidth}{%
    \fontsize{8}{8}\selectfont
        \textbf{Prompt P4}
        \hrule 
\vspace{1em}
\textbf{Stage 1 }
\\
You are a legal expert specializing in insurance.
\\
Given the following insurance clause, provide a simple but thorough response for the provided question.

You should provide enough detail to allow subsequent decisions to be made from the response but only restate what is explicitly provided in the clause and make no inferences or assumptions. 
\\
\\
\#\#\#\# BEGIN QUESTION \#\#\#\#

         [Question]
\\
\#\#\#\# END QUESTION \#\#\#\#
\\
\\
\#\#\#\# BEGIN TEXT\#\#\#\#

    [Clause]
\\
\#\#\#\# END TEXT \#\#\#\#
\\
\\
\#\#\#\# BEGIN RESTRICTIONS \#\#\#\#
\\
Do not use any inferred or implied information, answer based solely on the clause text provided and the answer options. Be as precise as possible when reasoning about which answer options are applicable.
\\
\\
If the clause does not explicitly state what type of insurance(s) a party must have, respond with "The clause does not contain sufficient information". 
\\
\\ 
\#\#\#\# END RESTRICTIONS \#\#\#\#
\\
\\
Your response will be subsequently mapped to the following options, so be sure that your answer is thorough:
     
         [Options]

    }%
}  
\caption{The first stage of a more restrictive two stage prompt \textbf{P4} for question \textbf{Q3} "What type(s) of insurance is a party required to have?". The second stage is identical to that of prompt \textbf{P3}.}
\label{fig:twostage-prompts-complex-Q3} 
\vspace{-1.5em}
\end{figure*}
\begin{figure*}
 \noindent\fbox{%
    \parbox{\textwidth}{%
    \fontsize{8}{8}\selectfont
        \textbf{Prompt P4}
        \hrule 
\vspace{1em}
\textbf{Stage 1 }
\\
You are a legal expert specializing in Force Majeure.
\\
Given the following Force Majeure clause, provide a simple but thorough response for the provided question.

You should provide enough detail to allow subsequent decisions to be made from the response but only restate what is explicitly provided in the clause and make no inferences or assumptions. 
\\
\\
\#\#\#\# BEGIN QUESTION \#\#\#\#

         [Question]
\\
\#\#\#\# END QUESTION \#\#\#\#
\\
\\
\#\#\#\# BEGIN TEXT\#\#\#\#

    [Clause]
\\
\#\#\#\# END TEXT \#\#\#\#
\\
\\
\#\#\#\# BEGIN RESTRICTIONS \#\#\#\#
\\
Do not use any inferred or implied information, answer based solely on the clause text provided and the answer options. Be as precise as possible when reasoning about which answer options are applicable.
\\
\\
If the clause does not explicitly state what events are included in the force majeure events, respond with "The clause does not contain sufficient information". 
\\
\\
A force majeure clause in an agreement specifies the events and circumstances (such as acts of God, war, utility failures, public health events like epidemics and pandemics, labour disruptions, etc.) beyond the reasonable control of the parties for which the parties are not liable. 
\\
\\
When responding, note that the "utilities" is defined as: energy, power, fuel, internet, communications, network, and any other utilities. 
"Failures" is defined as: any shortages, failures, lack of, inability to obtain, restrictions, outages, blackouts, or similar. 
\\
\\
If the force majeure clause may also contain catch-all language that states that the party is not liable or obligated to fulfill the terms of the agreement for any events beyond their reasonable control, include "There is catch-all language in the clause". in the response.  
\\
\\
\#\#\#\# END RESTRICTIONS \#\#\#\#
\\
\\
Your response will be subsequently mapped to the following options, so be sure that your answer is thorough:
     
         [Options]

    }%
}  
\caption{The first stage of a more restrictive two stage prompt \textbf{P4} for question \textbf{Q4} "What are the force majeure trigger events?". The second stage is identical to that of prompt \textbf{P3}.}
\label{fig:twostage-prompts-complex-Q4} 
\vspace{-1.5em}
\end{figure*}
\begin{figure*}
 \noindent\fbox{%
    \parbox{\textwidth}{%
    \fontsize{6}{6}\selectfont
        \textbf{Prompt P5}
        \hrule 
\vspace{1em}
\textbf{Stage 1 }
\\
You are a legal expert specializing in Force Majeure.
\\
Given the following  Force Majeure clause, provide a simple but thorough response for the provided question.

You should provide enough detail to allow subsequent decisions to be made from the response but only restate what is explicitly provided in the clause and make no inferences or assumptions.  
\\
\\
\#\#\#\# BEGIN QUESTION \#\#\#\#

         [Question]
\\
\#\#\#\# END QUESTION \#\#\#\#
\\
\\
\#\#\#\# BEGIN TEXT\#\#\#\#

    [Clause]
\\
\#\#\#\# END TEXT \#\#\#\#
\\
\\
\#\#\#\# BEGIN RESTRICTIONS \#\#\#\#
\\
\\
Do not use any inferred or implied information, answer based solely on the clause text provided and the answer options. Be as precise as possible when reasoning about which answer options are applicable.
\\
\\
If the clause does not explicitly state what events are included in the force majeure events, respond with "The clause does not contain sufficient information".  
\\
\\
A force majeure clause in an agreement specifies the events and circumstances (such as acts of God, war, utility failures, public health events like epidemics and pandemics, labour disruptions, etc.) beyond the reasonable control of the parties for which the parties are not liable. 
\\
\\
When responding, that utility or infrastructure failures include (but not exhaustively) the following events:  
\\- shortages of fuel, raw materials, power or energy, 
\\- power outages, blackouts, or electrical outages, power surges,
\\- failures or breakdowns of communication lines, Internet, 
\\- breakdown or failure of transmission, communication or computer facilities,
\\- scarcity or rationing of gasoline or other fuel or vital products,
\\- delay or inaccuracy in the transmission or reporting of orders due to a breakdown or failure of computer services, transmission or communication facilities,
\\- mechanical and/or electrical breakdown, 
\\- network catastrophes,
\\- shortages of fuel,
\\- hindrance in obtaining energy,
\\- restriction on the use of power
\\
\\
If the force majeure clause may also contain catch-all language that states that the party is not liable or obligated to fulfill the terms of the agreement for any events beyond their reasonable control, include "There is catch-all language in the clause." in the response.
\\
\\
\#\#\#\# END RESTRICTIONS \#\#\#\#
        \hrule 
\vspace{1em}
\textbf{Stage 2}
\\
\\
You are a legal expert seeking to answer a specific multiple choice question about  Force Majeure.

The following is an answer to the question: "[Question]". 
\\
\\
\#\#\#\# BEGIN RESPONSE\#\#\#\#

    [Response]
\\
\#\#\#\# END RESPONSE \#\#\#\#
\\
\\
Based upon this response, which of the following apply as the answer to the question "[Question]":
\\
\#\#\#\# Answer Options \#\#\#\#
     
         [Options]
\\
\#\#\#\# Answer Options \#\#\#\#
\\
\\
\#\#\#\# BEGIN RESTRICTIONS \#\#\#\#
\\
\\
You may select more than one option but you should be certain the answer option is correct if you do so. Prefer being precise over being inclusive.
\\
\\
Use appropriate legal definitions as necessary. You may assume this is in a US jurisdiction.  
\\
\\
Do not use any inferred or implied information, answer based solely on the clause text provided and the answer options. Be as precise as possible when reasoning about which answer options are applicable.
\\
\\
A force majeure clause in an agreement specifies the events and circumstances (such as acts of God, war, utility failures, public health events like epidemics and pandemics, labour disruptions, etc.) beyond the reasonable control of the parties for which the parties are not liable. 
\\
\\
When responding, note that Utility or infrastructure failures include (but not exhaustively) the following events:  
\\- shortages of fuel, raw materials, power or energy, 
\\- power outages, blackouts, or electrical outages, power surges,
\\- failures or breakdowns of communication lines, Internet, 
\\- breakdown or failure of transmission, communication or computer facilities,
\\- scarcity or rationing of gasoline or other fuel or vital products,
\\- delay or inaccuracy in the transmission or reporting of orders due to a breakdown or failure of computer services, transmission or communication facilities,
\\- mechanical and/or electrical breakdown, 
\\- network catastrophes,
\\- shortages of fuel,
\\- hindrance in obtaining energy,
\\- restriction on the use of power
\\Include option (d) in your response for any utility failures as defined above.
\\
\\
If the response states that there is catch-all langauge in the clause, include "f) There is catch-all language in the clause" in the response.
\\
\\
If the clause does not explicitly state what events are included in the force majeure events, respond with: "Unable to determine". 
\\
\\
\#\#\#\# END RESTRICTIONS \#\#\#\#
\\
\\
Return your response as a JSON array with each element taking the following format:
\\
```json
\\
\\
\{
\\
\\
    "bucket" : character  // Respond with the character representing the appropriate bucket.
\\
\}
\\
```
    }%
}  
\caption{Prompt \textbf{P4} modified to include an exhaustive definition of utility failures for question \textbf{Q4} "What are the force majeure trigger events?"}
\label{fig:twostage-prompts-complex-force-majeure} 
\vspace{-1.5em}
\end{figure*}

\begin{figure*}
    \centering
    \includegraphics[scale=0.6]{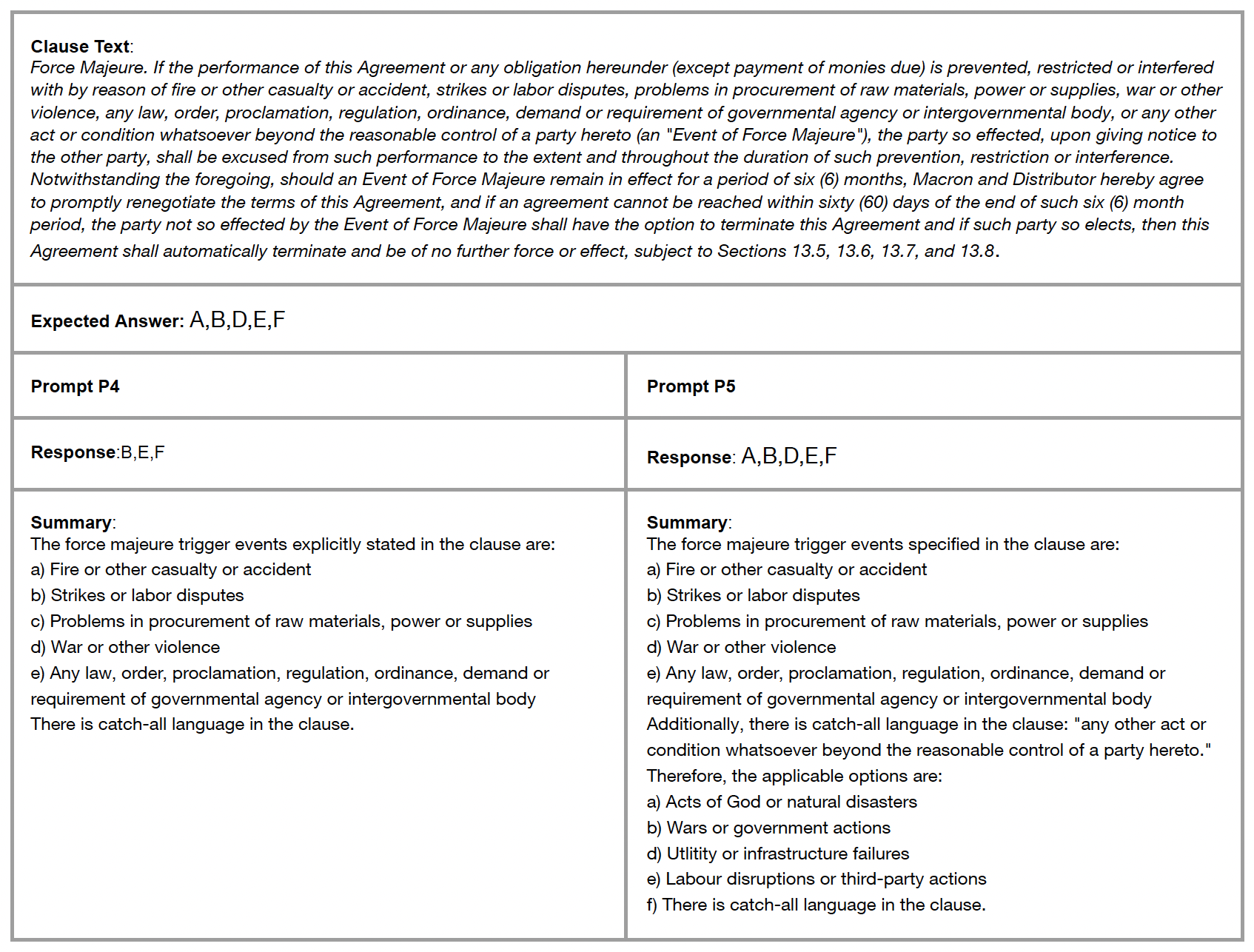}
    \caption{Responses of the GPT-4o model on a sample force majeure clause for the question \textbf{Q4} "What are the force majeure trigger events?"}
    \label{fig:force-majeure-analysis}
\end{figure*}
\end{document}